\documentclass[10pt,twocolumn,letterpaper]{article}

\usepackage{cvpr}              
\usepackage{colortbl}
\usepackage{xcolor}
\usepackage[accsupp]{axessibility}
\usepackage{makecell}
\usepackage{bm}
\usepackage{multirow}
\usepackage{xcolor}
\definecolor{color3}{RGB}{255, 255, 200}
\definecolor{color2}{RGB}{255, 220, 200}
\definecolor{color1}{RGB}{255, 181, 163}

\newcommand{\model}{\textit{IRGS}}

\definecolor{cvprblue}{rgb}{0.21,0.49,0.74}
\usepackage[pagebackref,breaklinks,colorlinks,allcolors=cvprblue]{hyperref}

\def\paperID{2510} 
\def\confName{CVPR}
\def\confYear{2025}

\title{IRGS: Inter-Reflective Gaussian Splatting with 2D Gaussian Ray Tracing}

\author{%
  Chun Gu$^{1,2}$
  \quad 
  Xiaofei Wei$^{1,2}$
  \quad 
  Zixuan Zeng$^1$
  \quad 
  Yuxuan Yao$^1$
  \quad
  Li Zhang$^{1,2}$\thanks{Corresponding author (\url{lizhangfd@fudan.edu.cn}).}
  \\
  $^1$School of Data Science, Fudan University 
  \quad
  $^2$Shanghai Innovation Institute
  \vspace{.5em} 
  \\
  \textcolor{black}{\url{https://fudan-zvg.github.io/IRGS}}
}

\begin{document}

\maketitle

\begin{abstract}

In inverse rendering, accurately modeling visibility and indirect radiance for incident light is essential for capturing secondary effects. Due to the absence of a powerful Gaussian ray tracer, previous 3DGS-based methods have either adopted a simplified rendering equation or used learnable parameters to approximate incident light, resulting in inaccurate material and lighting estimations. To this end, we introduce {inter-reflective Gaussian splatting} (\textbf{\model{}}) for inverse rendering. To capture inter-reflection, we apply the full rendering equation without simplification and compute incident radiance on the fly using the proposed differentiable \textbf{2D Gaussian ray tracing}. Additionally, we present an efficient optimization scheme to handle the computational demands of Monte Carlo sampling for rendering equation evaluation. Furthermore, we introduce a novel strategy for querying the indirect radiance of incident light when relighting the optimized scenes. Extensive experiments on multiple standard benchmarks validate the effectiveness of \model{}, demonstrating its capability to accurately model complex inter-reflection effects.

\end{abstract}    
\section{Introduction}
\label{sec:intro}

Inverse rendering is a fundamental problem in computer vision and graphics, aiming to reconstruct geometry and estimate material and lighting from a set of posed images. The introduction of neural radiance field (NeRF)~\cite{mildenhall2021nerf}, which uses a neural implicit field modeled by a multi-layer perceptron (MLP) to represent 3D scenes, has inspired many NeRF-based methods~\cite{srinivasan2021nerv,boss2021nerd,zhang2023neilf++,liu2023nero,jin2023tensoir} to address inverse rendering tasks. Leveraging NeRF's ray-based rendering technique, these methods can accurately compute the visibility and indirect radiance for incident light as required by the rendering equation. However, the exhaustive querying of a neural field during ray marching and its limited representational capacity constrain the efficiency and performance of neural implicit fields in inverse rendering.

\begin{figure}[t]
\centering
\includegraphics[width=0.94\linewidth]{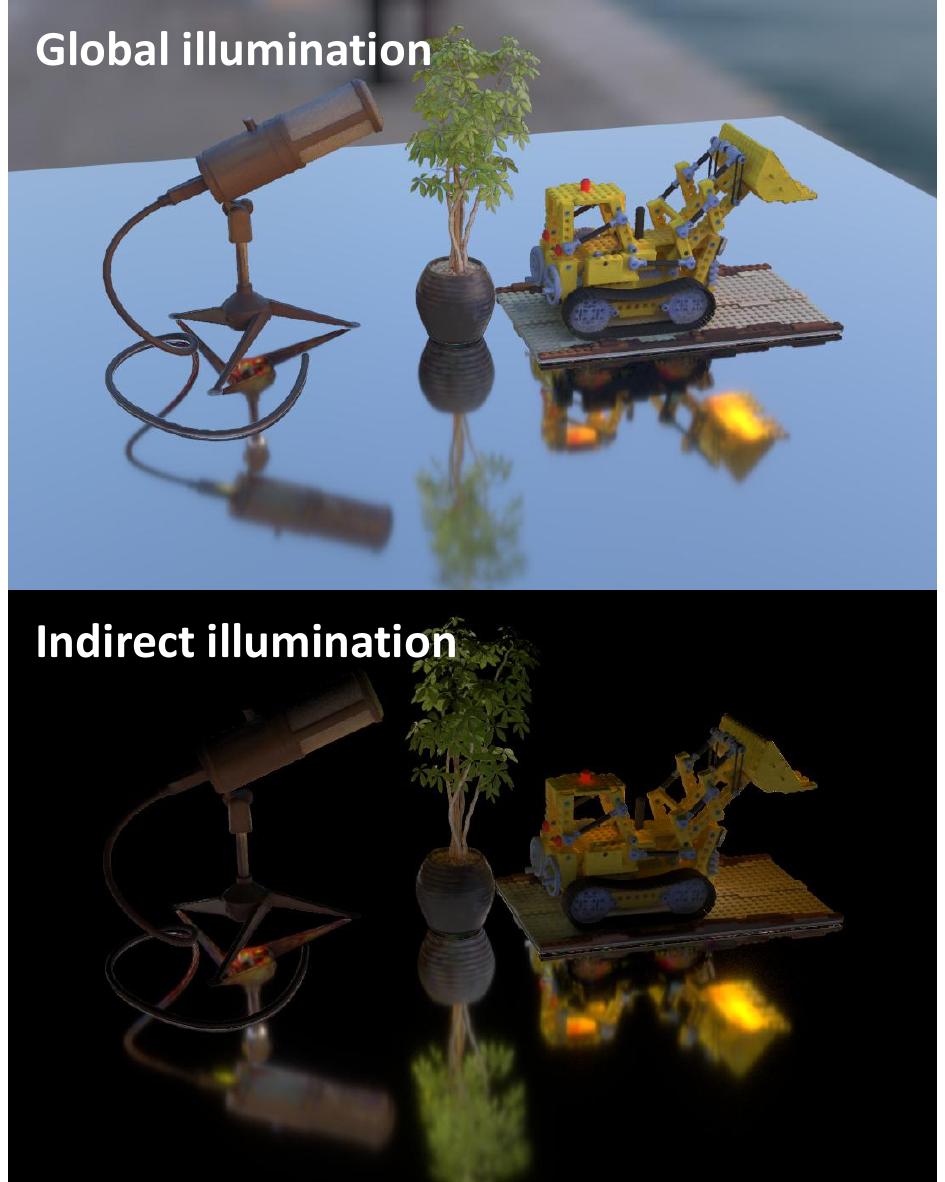}
\caption{
Global and indirect illumination in a Gaussian-based scene using our \model{}, highlighting its inter-reflective capabilities.
}
\vspace{-4mm}
\label{fig:teaser}
\end{figure}

Recently, 3D Gaussian splatting (3DGS)~\cite{kerbl3Dgaussians} has emerged as a promising technique for 3D scene representation, offering both remarkable rendering quality and efficiency. By modeling a static 3D scene as a collection of 3D Gaussians, 3DGS achieves real-time, high-quality rendering with a tile-based rasterizer. However, despite its strengths, 3DGS has notable limitations due to its reliance on rasterization, which limits its ability to accurately simulate ray-based effects that are essential for inverse rendering. As a result, some 3DGS-based inverse rendering methods~\cite{liang2024gs,shi2023gir,Jiang2023GaussianShader3G} have adopted simplified versions of the rendering equation, such as the split-sum approximation~\cite{munkberg2022extracting}. Other approaches, such as R3DG~\cite{gao2023relightable}, use learnable parameters to approximate indirect radiance, which fails to capture actual inter-reflection. These methods struggle to accurately model the rendering equation, limiting their effectiveness in realistic material and lighting estimation.

Our goal is to integrate the full rendering equation into the efficient Gaussian splatting for accurate inverse rendering, enabling the capture of complex interreflection effects. This requires efficient and precise computation of visibility and indirect radiance for incident light. Drawing inspiration from 3DGRT~\cite{MonneLoccoz20243DGR}, which achieves efficient ray tracing on 3D Gaussian primitives via a $k$-buffer hits-based marching technique and hardware-accelerated NVIDIA OptiX interface~\cite{parker2010optix}, we explore adapting this technique to model inter-reflection within the Gaussian splatting framework.

In this paper, we introduce \textbf{inter-reflective Gaussian splatting} (\textbf{\model{}}), a novel inverse rendering framework that leverages the proposed 2D Gaussian ray tracing technique to model inter-reflections within Gaussian splatting. To the best of our knowledge, this is the first method to integrate the full rendering equation without simplification into Gaussian splatting, accurately capturing inter-reflection effects by querying incident lighting through ray tracing. 
First, we propose \textbf{2D Gaussian ray tracing} to efficiently and accurately trace across 2D Gaussian primitives, computing the accumulated opacity and color of arbitrary rays. Unlike 3DGRT~\cite{MonneLoccoz20243DGR}, our approach features well-defined ray-splat intersections, enabling direct ray tracing on pretrained 2D Gaussian splatting~\cite{huang20242d} with minimal quality degradation. Next, we introduce the inter-reflective Gaussian splatting framework: starting from a pretrained 2D Gaussian splatting model, we apply the full rendering equation without simplification at intersection points determined by the depth map, while visibility and indirect radiance are computed on the fly using our 2D Gaussian ray tracing technique.
Notably, our 2D Gaussian ray tracing is fully differentiable, enabling the optimization of indirect light through backpropagation. We further propose an efficient optimization scheme to manage the computational demands of the exhaustive Monte Carlo sampling for the rendering equation. Additionally, we introduce a novel strategy to query indirect radiance during relighting, overcoming limitations in previous methods that often omit this component.
Extensive experiments on multiple standard benchmarks confirm the effectiveness of \model{}. In \cref{fig:teaser}, we visualize the global illumination (considering both direct and indirect radiance) and the indirect illumination (considering only indirect radiance) of a Gaussian-based scene rendered using \model{}, demonstrating that \model{} is capable of modeling remarkable inter-reflection effects.

The contributions of this paper are as follows:
\textbf{(i)} 2D Gaussian ray tracing, a technique that enables direct ray tracing on pretrained 2D Gaussian splatting with minimal quality degradation;
\textbf{(ii)} \model{}, an inverse rendering framework that incorporates the full rendering equation without simplification, using 2D Gaussian ray tracing to compute incident radiance on the fly;
\textbf{(iii)} An optimization scheme that manages computational demands for rendering equation evaluation;
\textbf{(iv)} A novel strategy that enables querying of indirect radiance during relighting.

\section{Related work}

\paragraph{Novel view synthesis} 
aims to leverage a limited collection of observed images of a scene to generate new images from an unseen viewpoint. Neural radiance field (NeRF)~\cite{mildenhall2021nerf} produces photo-realistic quality images by representing the scenes in terms of a volumetric radiance field encoded in a coordinate-based neural network. Subsequent advancements have focused on accelerating rendering\cite{mueller2022instant, reiser2021kilonerf}, enhancing image quality~\cite{barron2021mip,barron2022mip,barron2023zip}, and improving geometry reconstruction~\cite{li2023neuralangelo,wang2021neus,yariv2021volume}. More recently, 3D Gaussian splatting (3DGS)~\cite{kerbl3Dgaussians} introduced the use of fuzzy, anisotropic 3D Gaussian point cloud for scene representation along with an efficient tile-based rasterizer, achieving state-of-the-art results in both rendering quality and efficiency.
3DGS has inspired numerous approaches targeting various goals, including geometry reconstruction~\cite{huang20242d,Yu2024GOF}, dynamic scene reconstruction~\cite{yang2023real,yang2024deformable}, inverse rendering~\cite{gao2023relightable,liang2024gs}, and street scene applications~\cite{yan2024street,chen2023periodic}. However, the dependence on rasterization limits Gaussian splatting's ability to simulate ray-based effects accurately. To address this, 3DGRT~\cite{MonneLoccoz20243DGR} proposes a differentiable Gaussian ray tracer that efficiently performs ray tracing across 3D Gaussian primitives, allowing radiance computation along arbitrary rays. In this paper, we propose ray tracing on 2D Gaussian primitives. Unlike 3DGRT, 2D Gaussian primitives~\cite{huang20242d} have a well-defined ray-splat intersection, enabling direct ray tracing on pre-trained 2D Gaussian splatting with minimal quality degradation.

\begin{figure*}[ht]
\centering
\includegraphics[width=\linewidth]{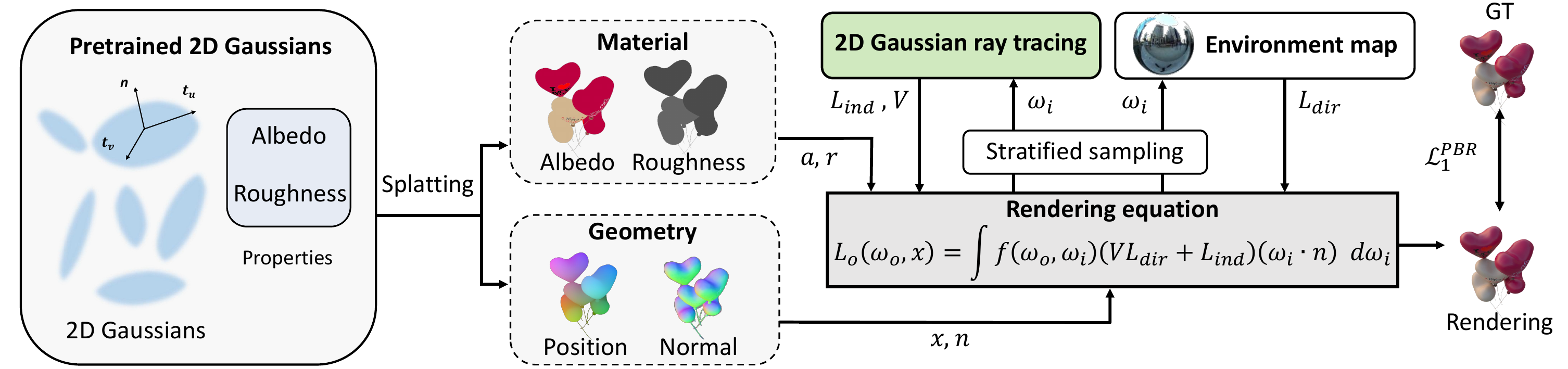}
\caption{
\textbf{Schematic illustration of the proposed IRGS.} Starting from a set of 2D Gaussians equipped with material properties, we apply rasterization to generate albedo, roughness, position, and normal maps. We then evaluate the rendering equation using stratified sampling at the corresponding position, drawing geometry and material values from these feature maps. The radiance of incident light is decomposed into direct radiance $L_\mathrm{dir}$ from the environment map, and indirect radiance $L_\mathrm{ind}$ with visibility $V$, obtained via 2D Gaussian ray tracing.
}
\vspace{-3mm}
\label{fig:pipeline}
\end{figure*}

\paragraph{Inverse rendering} 
focuses on recovering geometry, material, and lighting properties from captured images, presenting a complex challenge due to high-dimensional light-surface interactions. Inspired by the success of neural radiance fields, many methods~\cite{srinivasan2021nerv,boss2021nerd,yao2022neilf,zhang2023neilf++,boss2022samurai,attal2025flash,liu2023nero,jin2023tensoir,yang2023sireir,zhang2022modeling} leverage flexible ray marching techniques and neural implicit fields for inverse rendering.
Recent approaches~\cite{gao2023relightable,liang2024gs,Jiang2023GaussianShader3G,shi2023gir,wu2024deferredgs,guo2024prtgs,dihlmann2024subsurface,yao2025refGS} have adapted 3D Gaussian splatting (3DGS) to inverse rendering by assigning material-related properties to each Gaussian primitive. Leveraging the rapid rendering capabilities of 3DGS, these methods achieve much higher efficiency compared to NeRF-based techniques. However, due to the lack of an efficient Gaussian ray tracer, existing 3DGS-based methods either apply a simplified version of the rendering equation~\cite{liang2024gs} or use learnable parameters to model incident light~\cite{gao2023relightable}, which limits the accuracy of material and lighting estimation. For example, GS-IR~\cite{liang2024gs} employs split-sum approximation to simplify the rendering equation and uses baked volumes to store occlusion and indirect illumination. R3DG~\cite{gao2023relightable} shades each Gaussian individually, using point-based ray tracing to bake occlusion and spherical harmonics to parameterize indirect radiance.
To overcome these limitations, we propose the \model{} framework that utilizes 2D Gaussian ray tracing to accurately model inter-reflections.
\section{Method}

In this section, we introduce our inter-reflective Gaussian splatting with 2D Gaussian ray tracing, designed to reconstruct geometry and estimate material and lighting from a set of posed RGB images. We begin by providing the necessary background on 3D Gaussian splatting and the rendering equation (Section~\ref{sec:preliminary}). We then present the 2D Gaussian ray tracing technique, which performs ray tracing on 2D Gaussian primitives (Section~\ref{sec:2dgrt}). Finally, we outline our \model{} framework, which leverages the proposed 2D Gaussian ray tracing technique to query visibility and indirect radiance for incident light (Section~\ref{sec:irgs}).

\subsection{Preliminary}
\label{sec:preliminary}

\paragraph{3D Gaussian splatting}
3D Gaussian splatting (3DGS) represents a 3D scene using 3D Gaussian primitives, each defined as an ellipsoid characterized by a 3D mean $\boldsymbol{\mu}$ and a covariance matrix $\Sigma$. The influence of a Gaussian at a spatial position $\boldsymbol{x}$ is expressed as: 
\begin{equation} 
G(\boldsymbol{x}) = \exp\left(-\frac{1}{2} (\boldsymbol{x} - \boldsymbol{\mu})^\top \Sigma^{-1} (\boldsymbol{x} - \boldsymbol{\mu})\right). 
\end{equation} 
Each Gaussian is also assigned an opacity $o$ and a view-dependent color $\boldsymbol{c}$ modeled by spherical harmonics (SH). The rendering process of 3DGS starts by projecting 3D Gaussian primitives onto the 2D image plane. This is achieved by first applying a viewing transformation $W$ and then performing a perspective projection. The covariance matrix of the resulting 2D Gaussians can be approximated as $\Sigma' = J W \Sigma W^\top J^\top$, where $J$ represents the Jacobian matrix of the perspective projection. The final image is rendered by alpha-blending these projected 2D Gaussians from front to back: 
\begin{equation} 
\mathcal{C} = \sum_{i=1}^{N} T_i \alpha_i \boldsymbol{c}_i, \quad T_i = \prod_{j=1}^{i-1} (1 - \alpha_j), 
\label{eq:alpha_blending}
\end{equation} 
where $\alpha_i$ is calculated by multiplying opacity $o_i$ with each Gaussian’s influence, determined by $\Sigma'$ and the pixel coordinate. The covariance matrix $\Sigma$ is parameterized by a unit quaternion and a scaling vector to facilitate optimization.

\paragraph{Rendering equation}
In physically-based rendering, illumination at a given surface point $\boldsymbol{x}$ is computed via the rendering equation:
\begin{equation}
L_o(\boldsymbol{\omega}_o, \boldsymbol{x}) = \int_{\Omega} f(\boldsymbol{\omega}_o, \boldsymbol{\omega}_i, \boldsymbol{x}) L_\mathrm{i}(\boldsymbol\omega_i, \boldsymbol{x}) (\boldsymbol\omega_i\cdot \mathbf{n}) d\boldsymbol\omega_i\,,
\label{eq:rendering_equation}
\end{equation}
where $\boldsymbol{\omega}_o$ is the outgoing direction, $\boldsymbol{\omega}_i$ is the incident direction over the hemisphere defined by the surface normal $\boldsymbol{n}$, $f$ is the bidirectional reflectance distribution function (BRDF), and $L_\mathrm{i}(\boldsymbol{\omega}_i, \boldsymbol{x})$ is the radiance of the incident light. In this work, we use a simplified Disney BRDF model~\cite{burley2012physically} with only diffuse albedo $\boldsymbol{a}$ and roughness $r$ as parameters, and assume the material is dielectric. The BRDF $f$ in \cref{eq:rendering_equation} can be separated into a diffuse term $f_d = \frac{\boldsymbol{a}}{\pi}$ and a specular term $f_s$: 
\begin{equation}
f_s(\boldsymbol{\omega}_i,\boldsymbol{\omega}_o, \boldsymbol{x}) = \frac{D F G}{4 (\boldsymbol{\omega}_i \cdot \mathbf{n}) (\boldsymbol{\omega}_o \cdot \mathbf{n})},
    \label{eq:f_s}
\end{equation}
where $D$, $F$, and $G$ represent the normal distribution function, the Fresnel term, and the geometry term, respectively.

\subsection{2D Gaussian ray tracing}
\label{sec:2dgrt}

While 3DGS~\cite{kerbl3Dgaussians} has been successful in applications such as 3D reconstruction, achieving superior rendering quality and efficiency compared to NeRF-based methods, it is limited by the constraints of rasterization. These limitations are primarily related to ray-based effects, including rendering with highly distorted cameras and handling secondary ray effects. To address these challenges, 3DGRT~\cite{MonneLoccoz20243DGR} proposes performing ray tracing across 3D Gaussian primitives. This approach overcomes the inefficiencies of ray tracing on 3D Gaussian primitives by implementing a $k$-buffer hits-based marching technique with a hardware-accelerated NVIDIA OptiX interface~\cite{parker2010optix}. In this paper, as our focus is on inverse rendering, performing ray tracing on Gaussian primitives is essential to accurately compute inter-reflection effects when evaluating the rendering equation, including visibility and indirect radiance of incident light.

\begin{figure}[t]
\centering
\includegraphics[width=\linewidth]{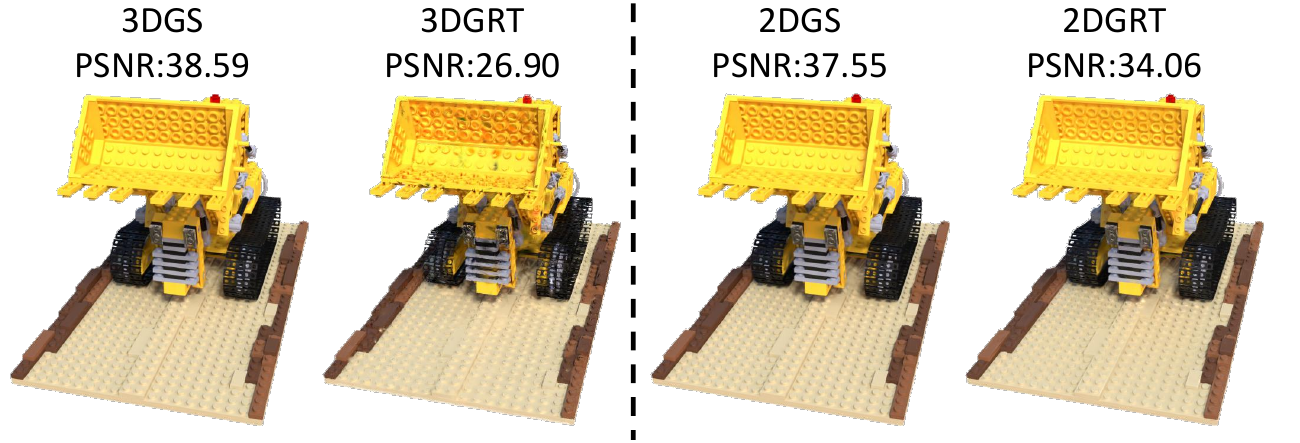}
\caption{
Performance of directly applying Gaussian ray tracing on a pretrained Gaussian splatting checkpoint in 3D Gaussian and 2D Gaussian cases, respectively. Our proposed 2DGRT achieves significantly less quality degradation than 3DGRT~\cite{MonneLoccoz20243DGR} in both quantitative metrics and visual results.
}
\vspace{-5mm}
\label{fig:raytracing-splatting}
\end{figure}

3D Gaussian ray tracing~\cite{MonneLoccoz20243DGR} performs ray tracing on 3D Gaussian primitives and determines the ray-splat intersection as the point of maximum response from the corresponding 3D Gaussian along the ray. This behavior differs from that of 3D Gaussian splatting, which computes the particle response in 2D image space, leaving the position of the 3D intersection point undefined. To leverage both the efficiency of Gaussian splatting and the flexible rendering capabilities of Gaussian ray tracing, we aim to perform ray tracing directly on the Gaussian splatting model for incident radiance inference. However, as shown in \cref{fig:raytracing-splatting}, applying ray tracing directly on a 3DGS checkpoint results in a noticeable decline in rendering quality compared to splatting. To address this issue, we propose \textbf{2D Gaussian ray tracing} (2DGRT), which performs ray tracing on 2D Gaussian primitives~\cite{huang20242d}. Since 2D Gaussian disks have a well-defined ray-splat intersection, this approach eliminates inconsistencies in intersection points present in 3D Gaussians.

Given a 2D Gaussian~\cite{huang20242d} defined by its center point $\boldsymbol{\mu}\in\mathbb{R}^3$, an opacity parameter $o\in [0, 1]$, two principal tangential vectors $\boldsymbol{t_u}\in\mathbb{R}^3$ and $\boldsymbol{t_v}\in\mathbb{R}^3$, and a scaling vector $\boldsymbol{s} = (s_u, s_v) \in\mathbb{R}^2$, we follow 3DGRT~\cite{MonneLoccoz20243DGR} to fit a bounding proxy of each 2D Gaussian using an adaptive icosahedron mesh, ensuring that a minimum influence of $\alpha_\mathrm{min}$ is enclosed. The vertices $\boldsymbol{v}$ of the bounding icosahedron mesh are obtained by transforming a regular icosahedron with a unit inner sphere:
\begin{equation} 
\boldsymbol{v} \gets \sqrt{2 \log(o / \alpha_\mathrm{min})} 
\begin{pmatrix} 
s_u \boldsymbol{t_u} & s_v \boldsymbol{t_v} & \epsilon\mathbf{1} 
\end{pmatrix} 
\boldsymbol{v} + \boldsymbol{\mu}, 
  \label{eq:icosahedron_transform}
\end{equation}
where $\epsilon$ denotes a small positive number.
Each icosahedron mesh consists of 20 triangular faces, resulting in a total of $20N$ triangles for $N$ 2D Gaussian primitives. To leverage the hardware-accelerated ray-triangle intersection provided by OptiX~\cite{parker2010optix}, we construct a Bounding Volume Hierarchy (BVH) using these triangles. By applying the $k$-buffer per-ray sorting algorithm~\cite{MonneLoccoz20243DGR}, we obtain the exact ordering of intersected Gaussians for each ray. To ensure consistency with 2DGS~\cite{huang20242d}, we analytically compute the ray-splat intersection point $\boldsymbol{p}$:
\begin{equation} 
\boldsymbol{p} = \boldsymbol{r}_o + \tau \boldsymbol{r}_d, \quad \text{where} \quad \tau = \frac{\boldsymbol{n}^\top(\boldsymbol{\mu} - \boldsymbol{r}_o)}{\boldsymbol{n}^\top\boldsymbol{r}_d}.
  \label{eq:raytracing_intersection}
\end{equation}
where $\boldsymbol{n} = \boldsymbol{t}_u \times \boldsymbol{t}_v$ is the normal vector of the 2D Gaussian disk, and $\boldsymbol{r}_o$ and $\boldsymbol{r}_d$ represent the origin and direction of the given ray, respectively. Thus, the influence of the current 2D Gaussian primitive is given by:
\begin{equation}
    G(\boldsymbol{p}) = \exp\left(-\frac{u^2 + v^2}{2}\right),
\label{eq:2d_gaussian}
\end{equation}
where $u = \frac{1}{s_u} (\boldsymbol{p} - \boldsymbol{\mu})^\top \boldsymbol{t}_u$, $v = \frac{1}{s_v} (\boldsymbol{p} - \boldsymbol{\mu})^\top \boldsymbol{t}_v$.

Consequently, the RGB value and accumulated opacity value of the ray can be obtained by alpha-blending (\cref{eq:alpha_blending}) the intersected 2D Gaussians from front to back: 
\begin{equation}
    (\boldsymbol{c}_\mathrm{rt}, o_\mathrm{rt}) \leftarrow\mathrm{Trace}(\boldsymbol{r}_o, \boldsymbol{r}_d),
    \label{eq:trace}
\end{equation}
where the RGB value of each Gaussian is modeled by the view-dependent color $\boldsymbol{c}$ as in vanilla 2D Gaussian splatting.
It is worth noting that subtle differences remain between 2D Gaussian ray tracing and 2DGS due to sorting errors introduced by per-tile sorting in 2DGS. However, as shown in \cref{fig:raytracing-splatting}, the quality degradation from 2D Gaussian ray tracing is minimal and significantly less noticeable than that observed in 3D Gaussians.

\subsection{Inter-reflective Gaussian splatting}
\label{sec:irgs}

\subsubsection{Stage I: 2D Gaussian pretraining}
Before estimating material and lighting for inverse rendering, we first pretrain a standard 2D Gaussian splatting model. This step is essential, as reliable geometry is required for inverse rendering, and evaluating the rendering equation is computationally intensive. Using a pretrained 2D Gaussian splatting model can greatly accelerate the training process. Along with the RGB rendering described in \cref{eq:alpha_blending}, we also render normalized depth and normal maps as follows:
\begin{equation}
    \{\mathcal{D}, \mathcal{N}\} = \sum_{i=1}^{N} w_i\{d_{i},\boldsymbol{n}_{i}\},\,\text{where}\, w_i=\frac{T_i \alpha_i}{\sum_{i=1}^{N}T_i \alpha_i}.
    \label{eq:depth_normal}
\end{equation}

\noindent{\bf Training loss}
Following 2DGS~\cite{huang20242d}, we apply a normal consistency loss $\mathcal{L}_\mathrm{n}$ to ensure alignment between the rendered normal map and the underlying geometry, as well as a depth distortion loss~\cite{huang20242d} $\mathcal{L}_\mathrm{d}$ to encourage concentration of the 2D Gaussians. Additionally, we introduce an edge-aware smoothness loss~\cite{gao2023relightable,guo2022nerfren} $\mathcal{L}_\mathrm{s,n}$ on the normal map to enhance geometric robustness:
\begin{equation}
    \mathcal{L}_\mathrm{s,n} = \Vert \nabla \mathcal{N} \Vert \exp(-\Vert \nabla \mathcal{C}_\mathrm{gt} \Vert) ,
\label{eq:reg_smooth_normal}
\end{equation}
where $\mathcal{C}_\mathrm{gt}$ is the ground truth training image. 
To address the presence of floaters during standard 2DGS training, we also incorporate a binary cross-entropy loss~\cite{wang2021neus} to constrain the geometry using the provided object mask $\mathcal{M}$:
\begin{equation}
    \mathcal{L}_\mathrm{o} = -\mathcal{M}\log{\mathcal{O}} - (1-\mathcal{M})\log{(1-\mathcal{O})},
\end{equation}
where $\mathcal{O} = \sum_{i=1}^{N} T_i \alpha_i$, denotes the accumulated opacity map. Combined with the RGB reconstruction loss $\mathcal{L}_\mathrm{c}$ from 3DGS~\cite{kerbl3Dgaussians}, the training loss for the first stage is defined as:
\begin{equation}
    \mathcal{L}^\mathrm{1} = \mathcal{L}_\mathrm{c}+\lambda_\mathrm{n}\mathcal{L}_\mathrm{n}+\lambda_\mathrm{d}\mathcal{L}_\mathrm{d}+\lambda_\mathrm{s,n}\mathcal{L}_\mathrm{s,n}+\lambda_\mathrm{o}\mathcal{L}_\mathrm{o}.
\end{equation}

\subsubsection{Stage II: inverse rendering}

The order of shading and rasterization is crucial in physically based rendering with Gaussian splatting. Previous works, such as R3DG~\cite{gao2023relightable}, propose performing shading on each Gaussian, taking advantage of properties modeled on each Gaussian, such as indirect radiance. However, since the rendered normal map is supervised at the pixel level after rasterization, shading on individual Gaussians can lead to inaccurate and blurred results. Consequently, we propose applying the rendering equation after rasterization.

\noindent{\bf Rasterization}
As introduced in Section~\ref{sec:preliminary}, a dielectric BRDF model is used for the rendering equation. We assign each 2D Gaussian primitive additional material properties: diffuse albedo $\boldsymbol{a}$ and roughness $r$. Pixel-level albedo map $\mathcal{\boldsymbol{A}}$ and roughness map $\mathcal{R}$ are obtained through alpha-blending during rasterization: 
$\{\mathcal{\boldsymbol{A}}, \mathcal{R}\} = \sum_{i=1}^{N} w_i\{\boldsymbol{a}_{i},r_{i}\}$.

Previous inverse rendering methods using Gaussian splatting have lacked an efficient way to query the visibility and radiance for arbitrary incident directions. As a result, these methods typically apply simplified versions of the rendering equation~\cite{liang2024gs} (e.g., split-sum approximation~\cite{munkberg2022extracting}) or use learnable parameters to model indirect light~\cite{gao2023relightable}. In this paper, we leverage the proposed 2D Gaussian ray tracing technique to efficiently and accurately query inter-reflection effects on the fly, including the visibility and indirect radiance of incident light. This allows us to apply the full rendering equation without simplification.

\noindent{\bf Light parametrization}
We decompose the incident radiance at surface point $\boldsymbol{x}$ with direction $\boldsymbol{\omega}_i$ in \cref{eq:rendering_equation} into direct radiance from infinity and indirect radiance:
\begin{equation}
    L_\mathrm{i}(\boldsymbol{\omega}_i, \boldsymbol{x}) = V(\boldsymbol{\omega}_i, \boldsymbol{x})L_\mathrm{dir}(\boldsymbol{\omega}_i)+L_\mathrm{ind}(\boldsymbol{\omega}_i, \boldsymbol{x}),
\label{eq:lighting_decompose}
\end{equation}
where $L_\mathrm{dir}(\boldsymbol{\omega}_i)$ is modeled using an environment cubemap, and $V(\boldsymbol{\omega}_i, \boldsymbol{x})$ along with $L_\mathrm{ind}(\boldsymbol{\omega}_i, \boldsymbol{x})$ are queried using 2D Gaussian ray tracing: 
\begin{equation}
    (L_\mathrm{ind}(\boldsymbol{\omega}_i, \boldsymbol{x}), 1-V(\boldsymbol{\omega}_i, \boldsymbol{x}))\leftarrow\mathrm{Trace}(\boldsymbol{x}, \boldsymbol{\omega}_i)
\end{equation}
It is important to note that the RGB values used in ray tracing correspond to the view-dependent color $\boldsymbol{c}$ from the first stage. Additionally, our 2D Gaussian ray tracing approach is fully differentiable, allowing gradients to propagate through the ray tracing process to optimize the indirect radiance of the incident ray.

\begin{figure*}[ht]
\centering
\includegraphics[width=0.98\linewidth]{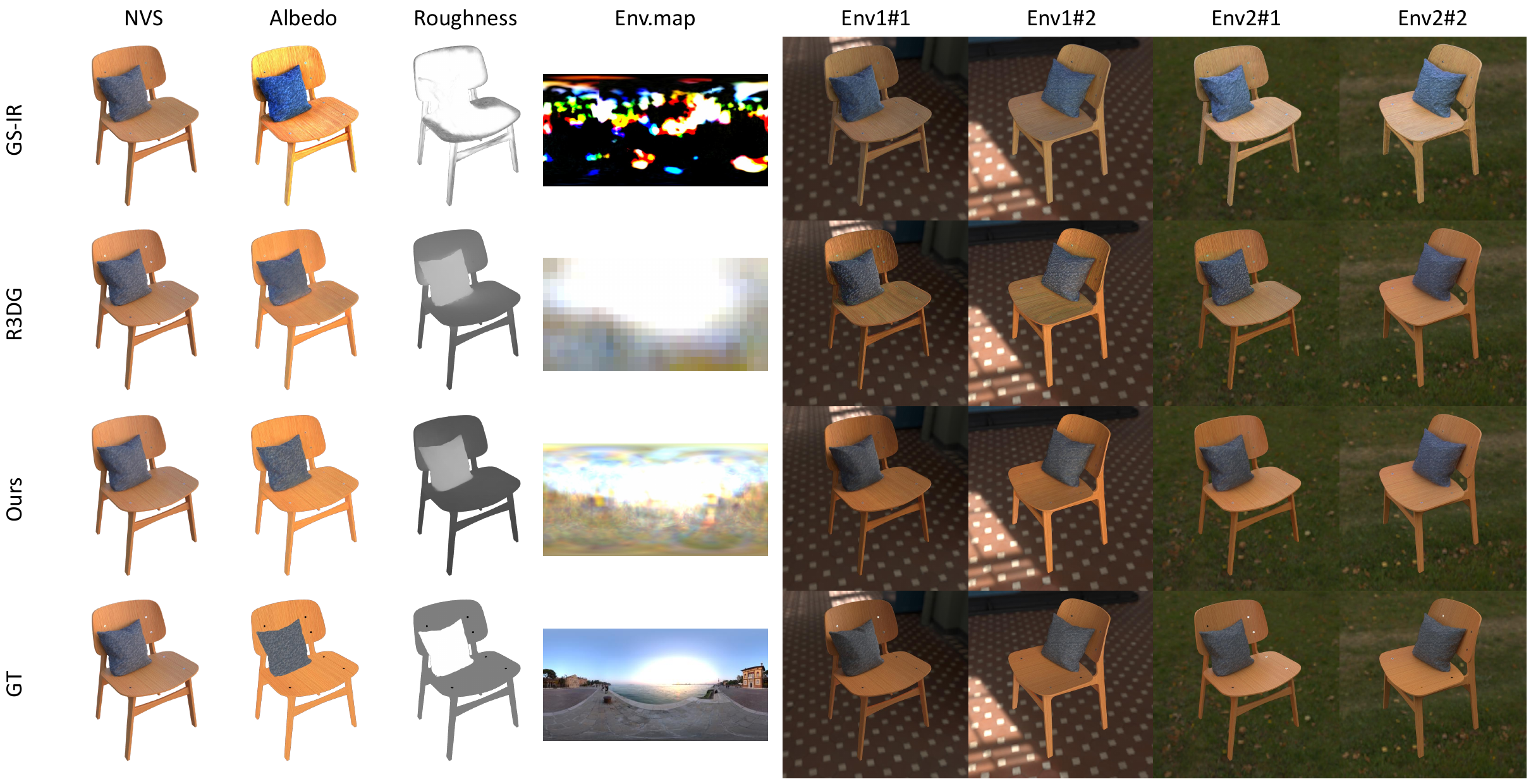}
\caption{
Qualitative comparison of NVS, material and lighting estimation, and relighting results on the Synthetic4Relight dataset~\cite{zhang2022modeling}.
}
\label{fig:exp_syn4_compare}
\end{figure*}
\begin{table*}[ht]
\centering
\caption{Quantitative comparison on the Synthetic4Relight dataset~\cite{zhang2022modeling}. A higher intensity of the red color signifies a better result.}
\resizebox{\linewidth}{!}{
\begin{tabular}{c|ccc|ccc|ccc|c|c}
\toprule[1.5px]
     & \multicolumn{3}{c|}{\textbf{Novel view synthesis}} & \multicolumn{3}{c|}{\textbf{Relighting}} & \multicolumn{3}{c|}{\textbf{Albedo}} & \textbf{Roughness} & Time \\
                  & PSNR↑  & SSIM↑ & LPIPS↓ & PSNR↑    & SSIM↑    & LPIPS↓   & PSNR↑   & SSIM↑  & LPIPS↓  & MSE↓  & (hours)    \\ \hline
NeRFactor~\cite{zhang2021nerfactor}         & 22.80  & 0.916 & 0.150  & 21.54    & 0.875    & 0.171    & 19.49   & 0.864  & 0.206   & N/A   & $>$48    \\
InvRender~\cite{zhang2022modeling}         & 30.74  & 0.953 & 0.086  & 28.67    & 0.950    & 0.091    & 28.28   & 0.935  & 0.072   & \cellcolor{red!60}0.008 & 14     \\
TensorIR~\cite{jin2023tensoir}          & \cellcolor{red!30}35.80  & \cellcolor{red!30}0.978 & \cellcolor{red!10}0.049  & \cellcolor{red!10}29.69    & \cellcolor{red!10}0.951    & \cellcolor{red!10}0.079    & \cellcolor{red!30}30.58   & \cellcolor{red!10}0.946  & \cellcolor{red!10}0.065   & 0.015  & 3   \\
GS-IR~\cite{liang2024gs}    & 33.95 & 0.965 & 0.057 & 25.40 & 0.924 & 0.083 & 19.48 & 0.896 & 0.117 &  \cellcolor{red!30}0.011 & \cellcolor{red!60}0.4\\ 
R3DG~\cite{gao2023relightable}    &   \cellcolor{red!60}36.80  &  \cellcolor{red!60}0.982   &   \cellcolor{red!60}0.028     &  \cellcolor{red!30}31.00       &    \cellcolor{red!30}0.964      &    \cellcolor{red!30}0.050  &  \cellcolor{red!10}28.31   &    \cellcolor{red!30}0.951    &    \cellcolor{red!30}0.058     &    \cellcolor{red!10}0.013   &  \cellcolor{red!10}0.9   \\ 
\hline
IRGS (Ours)    &   \cellcolor{red!10}35.48  &  \cellcolor{red!10}0.974  &   \cellcolor{red!30}0.043     &   \cellcolor{red!60}34.90     &  \cellcolor{red!60}0.969      & \cellcolor{red!60}0.048   & \cellcolor{red!60}30.81  &   \cellcolor{red!60}0.957   &   \cellcolor{red!60}0.055    &   \cellcolor{red!60}0.008   &  \cellcolor{red!30}0.7  \\ 
\bottomrule[1.5px]
\end{tabular}
}
\label{tab:exp_syn4}
\end{table*}

\noindent{\bf Rendering}
From the rendered depth and normal maps, we can determine the surface position $\boldsymbol{x}$ and the corresponding surface normal vector $\boldsymbol{n}$ for each pixel coordinate. Monte Carlo sampling is used to evaluate the rendering equation. Since we focus on reconstructing objects with dielectric materials, we employ stratified sampling to uniformly sample $N_\mathrm{r}$ incident directions across the hemisphere. The final rendering result is then computed as:
\begin{equation}
    \mathbf{c}_\mathrm{pbr} = \frac{2\pi}{N_\mathrm{r}}\sum_{i=1}^{N_\mathrm{r}} (f_\mathrm{d}+f_\mathrm{s}) L_\mathrm{i}(\boldsymbol\omega_i, \boldsymbol{x})(\boldsymbol\omega_i\cdot \mathbf{n}),
    \label{eq:pbr}
\end{equation}

\begin{figure*}[t]
\centering
\includegraphics[width=\linewidth]{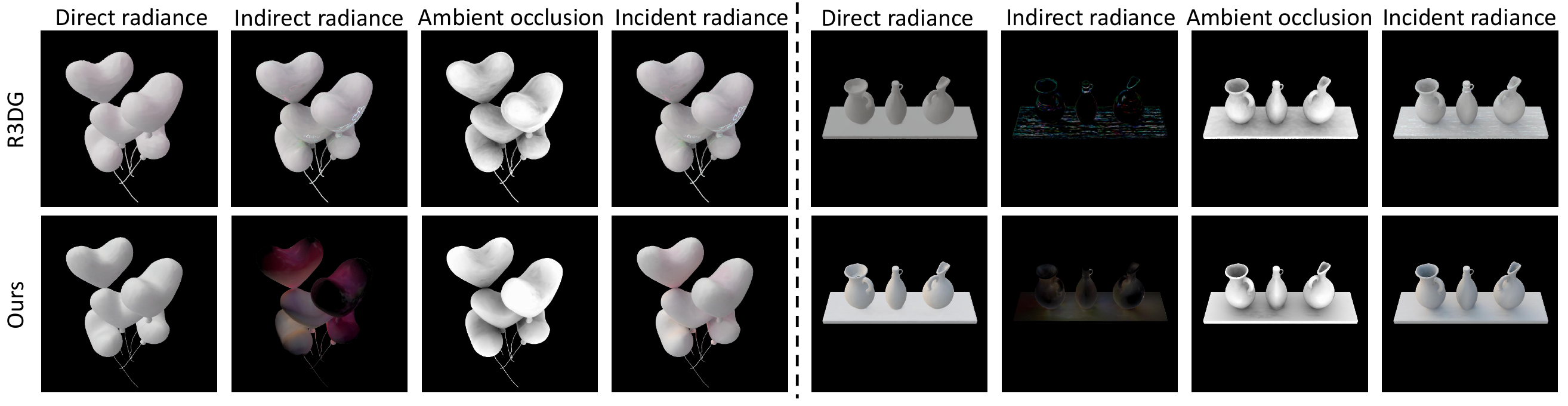}
\vspace{-7mm}
\caption{
Visualization of estimated components in incident light, including the averaged direct radiance $L_\mathrm{dir}$, indirect radiance $L_\mathrm{ind}$, visibility $V$ (ambient occlusion), incident radiance $L_\mathrm{i}$. Compared to R3DG~\cite{gao2023relightable}, \model{} achieves notably more realistic results, especially in estimated indirect light, due to its accurate modeling of inter-reflections.
}
\vspace{-2mm}
\label{fig:exp_material_light}
\end{figure*}

\noindent{\bf Training loss}
In addition to the training losses from the first stage, we apply regularization to facilitate material and lighting estimation. Given the prior that diffuse light $L_\mathrm{diffuse}=\frac{1}{N_\mathrm{r}}\sum_{i=1}^{N_\mathrm{r}}L(\boldsymbol{\omega}_i, \boldsymbol{x})$ exhibits a natural white radiance, we minimize the difference between the RGB values~\cite{liu2023nero}: 
\begin{equation}
    \mathcal{L}_\mathrm{light} = \sum_{c}(L_\mathrm{diffuse} - \frac{1}{3}\sum_{c}L_\mathrm{diffuse}), c\in\{R,G,B\}.
\label{eq:reg_light}
\end{equation}
Similar to \cref{eq:reg_smooth_normal}, we also apply edge-aware smoothness regularization $\mathcal{L}_\mathrm{s,a}$ and $\mathcal{L}_\mathrm{s,r}$ to the rendered albedo and roughness map, respectively, to ensure smooth material estimation.
Additionally, for training efficiency, we apply the rendering equation to only a subset of pixels in each iteration. We set a maximum of $N_\mathrm{rays}$ rays per iteration, meaning only $\left\lfloor N_\mathrm{rays}/N_\mathrm{r} \right\rfloor$ pixels are randomly selected for final rendering equation evaluation. This scheme allows us to use a large $N_\mathrm{r}$ when evaluating the rendering equation, significantly enhancing estimation quality. Consequently, only a L1 loss $\mathcal{L}_\mathrm{1}^\mathrm{pbr}$ is used to supervise the final color $\boldsymbol{c}_\mathrm{pbr}$. It is important to note that the rendered image $\mathcal{C}$ from the vanilla 2DGS is still optimized, while the view-dependent color $\boldsymbol{c}$ associated with each Gaussian is used to compute radiance in 2D Gaussian ray tracing. The total loss for the second stage is given by:
\begin{equation}
    \mathcal{L}^\mathrm{2} = \mathcal{L}^\mathrm{1}+\lambda_\mathrm{1}^\mathrm{pbr}\mathcal{L}_\mathrm{1}^\mathrm{pbr}+\lambda_\mathrm{light}\mathcal{L}_\mathrm{light}+\lambda_\mathrm{s,a}\mathcal{L}_\mathrm{s,a}+\lambda_\mathrm{s,r}\mathcal{L}_\mathrm{s,r},
\end{equation}

\noindent{\bf Relighting}
When performing relighting, changes in environmental lighting make the optimized radiance $\boldsymbol{c}$ of each Gaussian no longer applicable for indirect radiance in 2D Gaussian ray tracing. Unlike previous methods~\cite{gao2023relightable,liang2024gs} that omit indirect lighting during relighting, we propose using the split-sum approximation~\cite{munkberg2022extracting} to efficiently query indirect radiance without recursive evaluation. Specifically, we apply ray tracing on each incident ray to alpha-blend the albedo, roughness, and normal values, resulting in $\mathcal{\boldsymbol{A}}_\mathrm{r}$, $\mathcal{R}_\mathrm{r}$, and $\mathcal{\boldsymbol{N}}_\mathrm{r}$. These aggregated values are then shaded using a pre-filtered environment map, following the approach in Nvdiffrec~\cite{munkberg2022extracting}. Additionally, we apply importance sampling on the intensity of the ground truth environment map instead of stratified sampling, enabling more accurate rendering equation evaluation by concentrating samples in high-intensity directions. Please refer to the appendix for further details.

\section{Experiment}

\begin{figure*}[ht]
\centering
\includegraphics[width=0.97\linewidth]{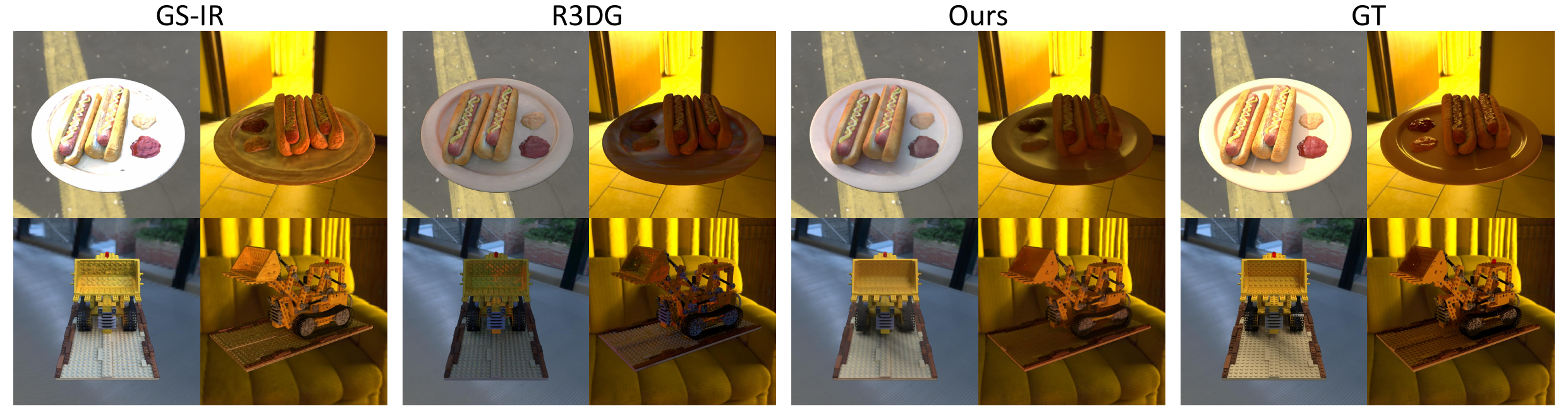}
\vspace{-3mm}
\caption{
Qualitative comparison of relighting results on TensoIR dataset~\cite{zhang2022modeling}.
}
\vspace{-4mm}
\label{fig:exp_tensoir_compare}
\end{figure*}

\noindent{\bf Datasets and metrics}
To validate the effectiveness of \model{}, we conduct extensive experiments on multiple datasets for the inverse rendering task, including two synthetic datasets: Synthetic4Relight dataset~\cite{zhang2022modeling} and the TensoIR dataset~\cite{jin2023tensoir}, as well as a real-world dataset: Stanford-ORB dataset~\cite{kuang2024stanford}. For quantitative comparisons, we follow previous works~\cite{gao2023relightable,liang2024gs} and employ various metrics to evaluate performance across different outputs. Specifically, we use PSNR, SSIM~\citep{wang2004image}, and LPIPS~\citep{zhang2018unreasonable} for assessing image quality in novel view synthesis, albedo, and relighting tasks. For normal and roughness, we use mean angular error (MAE) and  mean squared error (MSE), respectively.

\noindent{\bf Implementation details}
As detailed in Section~\ref{sec:irgs}, our training process consists of two stages. The first stage is trained over 40,000 iterations with the hyperparameters $\lambda_\mathrm{n}$, $\lambda_\mathrm{d}$, $\lambda_\mathrm{s,n}$, and $\lambda_\mathrm{o}$ set to $0.05$, $1000$, $0.02$, and $0.01$, respectively. The second stage continues for an additional 20,000 iterations, with $\lambda_\mathrm{1}^\mathrm{pbr}$, $\lambda_\mathrm{light}$, $\lambda_\mathrm{s,a}$, and $\lambda_\mathrm{s,r}$ set to $1.0$, $0.01$, $2.0$, and $2.0$, respectively. We implement 2D Gaussian ray tracing in OptiX~\cite{parker2010optix} via PyTorch CUDA extensions, updating the BVH at each training iteration for the varying geometry, with each update taking approximately $3$ ms.
In line with 3DGRT~\cite{MonneLoccoz20243DGR}, we use a buffer of $k = 16$ for per-ray sorting and terminate ray tracing when transmittance falls below $0.03$. We employ $N_\mathrm{r}=256$ rays for rendering equation evaluation, and each training iteration samples up to $N_\mathrm{rays} = 2^{18}$ rays. At $N_\mathrm{r}=256$, rendering a complete image takes about $1$ second. For environmental lighting, we use a $32 \times 32$ cubemap during optimization. The learning rates for albedo, roughness, and the cubemap are set to $0.005$, $0.005$, and $0.01$, respectively, with other hyperparameters following those in 2DGS~\cite{huang20242d}. The entire training process takes around 40 minutes, with 15 minutes for the first stage and 25 minutes for the second stage. All experiments are conducted on a single NVIDIA RTX 3090 GPU.

\begin{table}[!t]
\centering
\setlength\tabcolsep{10pt}
\caption{Quantatitive comparison on TensoIR dataset~\cite{jin2023tensoir}.}
\vspace{-3mm}
\scalebox{0.8}{
\begin{tabular}{c|c|c|c|c}
\Xhline{3\arrayrulewidth}
\multirow{2}{*}{Method} &
Normal & NVS & Albedo & Relight \\
 & MAE$\downarrow$ & PSNR$\uparrow$ & PSNR$\uparrow$ & PSNR$\uparrow$ \\
\hline
NeRFactor \cite{zhang2021nerfactor} &6.314 &
24.679 & 
25.125 &
23.383 \\
InvRender \cite{zhang2022modeling} &5.074 &
27.367 &
27.341 &
23.973 \\
TensoIR \cite{jin2023tensoir} &\cellcolor{red!60}4.100 &
35.088 & 
\cellcolor{red!10}29.275 & 
\cellcolor{red!30}28.580 \\
GS-IR~\cite{liang2024gs} & \cellcolor{red!10}4.948 &
\cellcolor{red!10}35.333 & 
\cellcolor{red!30}30.286 &
24.374 \\
R3DG~\cite{gao2023relightable} &5.927 &
 \cellcolor{red!60}37.343 &
 26.199 &
 \cellcolor{red!10}27.367\\
\hline
IRGS (Ours) &\cellcolor{red!30}4.112 &
 \cellcolor{red!30}35.876 &
 \cellcolor{red!60}33.796 & 
\cellcolor{red!60}29.907 \\
\Xhline{3\arrayrulewidth}
\end{tabular}
}
\vspace{-5mm}
\label{tab:exp_tensoir}
\end{table}

\subsection{Results}
\noindent{\bf Synthetic4Relight}
For experiments on Synthetic4Relight~\cite{zhang2022modeling}, we evaluate \model{}'s performance in novel view synthesis (NVS), relighting, material estimation, and efficiency. In \cref{tab:exp_syn4}, our approach achieves higher quality than previous methods across relighting, and material estimation, demonstrating \model{}'s ability to accurately capture material details. While R3DG~\cite{gao2023relightable} achieves higher performance in NVS, this is primarily due to its shading on Gaussians, which enhances NVS capability at the expense of relighting performance. \model{} also completes optimization in a relatively short time of $0.7$ hours, emphasizing its efficiency in inverse rendering applications. In \cref{fig:exp_syn4_compare}, we provide a qualitative comparison of the reconstructed “chair” scene against GS-based competitors~\cite{gao2023relightable,liang2024gs}, visualizing NVS, albedo, roughness, estimated environment maps, and relighting images. Our results demonstrate high-fidelity material estimation and realistic relighting outcomes. \cref{fig:exp_material_light} further illustrates a comparison of the estimated incident light components between \model{} and R3DG~\cite{gao2023relightable}, covering direct radiance, indirect radiance, ambient occlusion, and incident radiance. \model{} achieves more reasonable indirect radiance and clearer ambient occlusion, enhancing the overall realism.

\noindent{\bf TensoIR}
For TensoIR dataset~\cite{jin2023tensoir}, we report metrics for NVS, relighting, and albedo estimation in \cref{tab:exp_tensoir}. Compared to prior  arts~\cite{zhang2021nerfactor,zhang2022modeling,jin2023tensoir,liang2024gs,gao2023relightable}, our method consistently outperforms in relighting and albedo estimation, and achieving comparable performance in normal quality with TensoIR~\cite{jin2023tensoir}. 
Similar to Synthetic4Relight, R3DG~\cite{gao2023relightable} achieves higher accuracy in NVS. 
\cref{fig:exp_tensoir_compare} presents a qualitative comparison of relighting images generated by GS-based methods~\cite{liang2024gs,gao2023relightable}, showing that \model{} produces more realistic relighting results, with natural shadows.

\begin{figure}[t]
\centering
\includegraphics[width=0.97\linewidth]{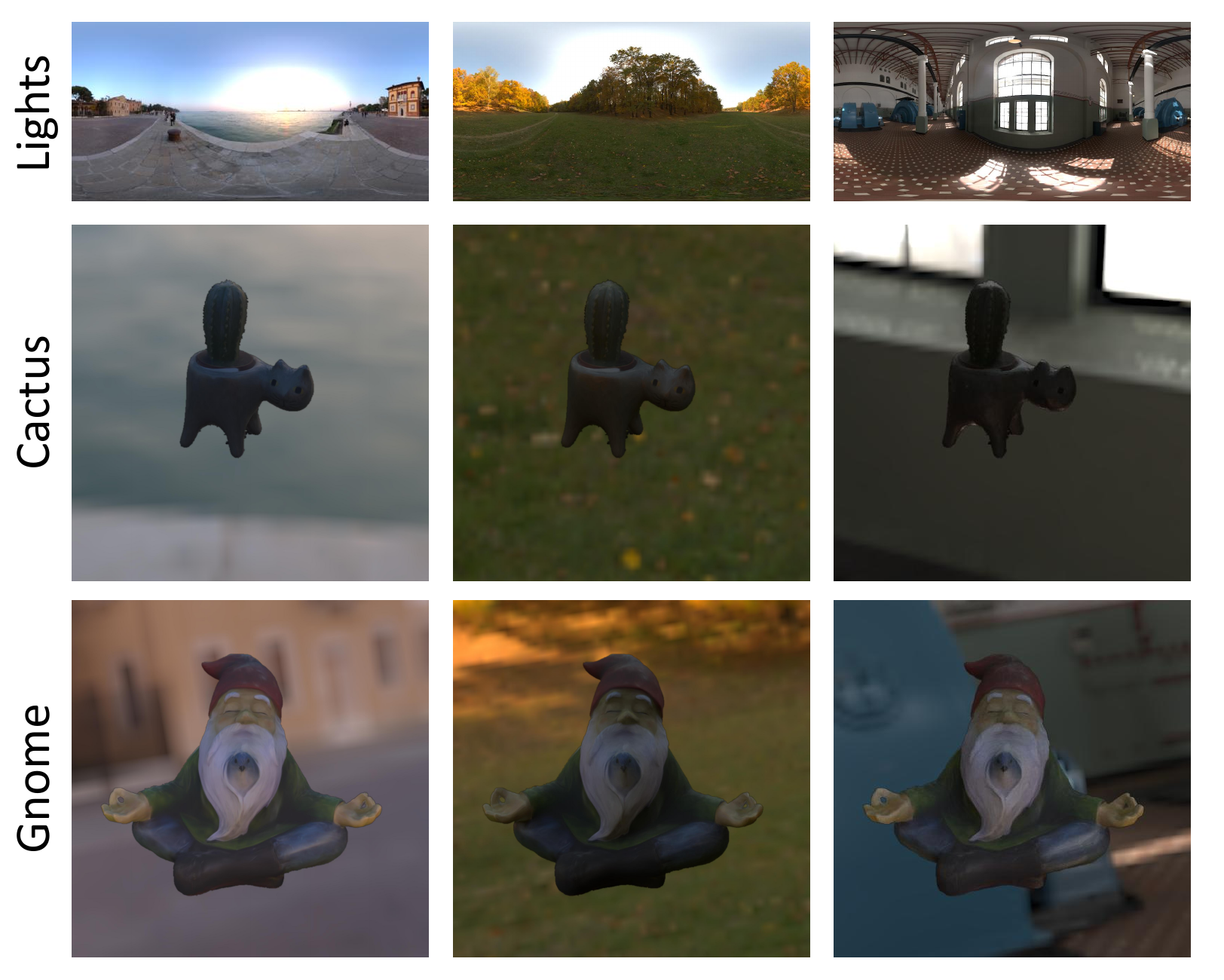}
\vspace{-4mm}
\caption{
Relighting results on Stanford-ORB dataset~\cite{kuang2024stanford}.
}
\vspace{-3mm}
\label{fig:exp_orb_relight}
\end{figure}

\noindent{\bf Stanford-ORB dataset}
We also conduct experiments on the real-world Stanford-ORB dataset~\cite{kuang2024stanford}. As shown in \cref{fig:exp_orb_relight}, \model{} produces realistic relighting results.

\begin{figure}[t]
\centering
\begin{subfigure}[t]{\linewidth}
    \centering
    \includegraphics[width=0.95\linewidth]{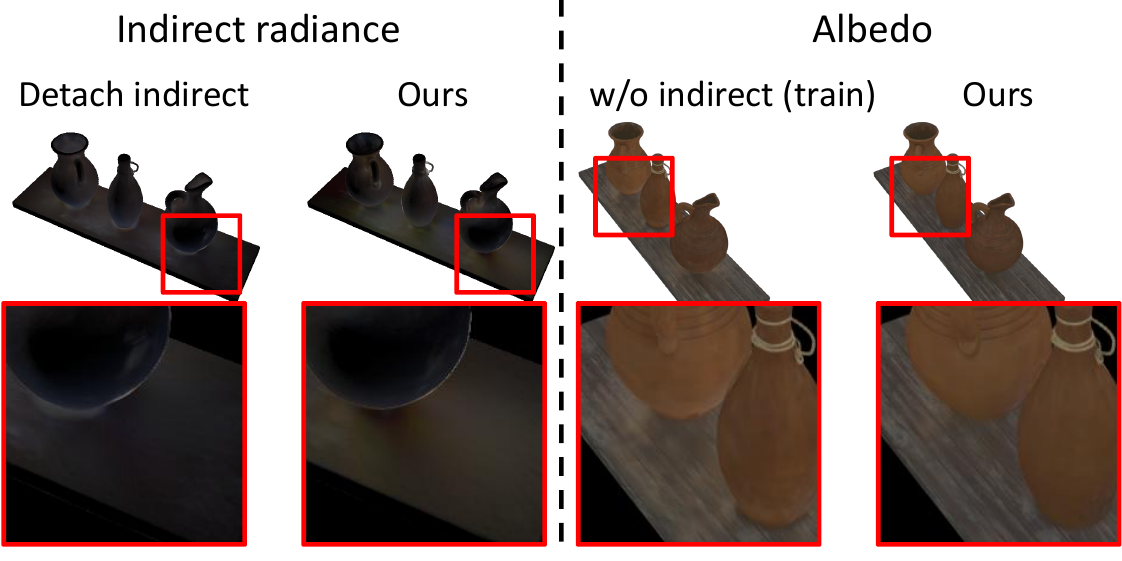}
    \caption{Ablation studies on indirect light modeling.}
    \label{fig:ablation_indirect}
\end{subfigure}
\begin{subfigure}[t]{\linewidth}
    \centering
    \includegraphics[width=0.95\linewidth]{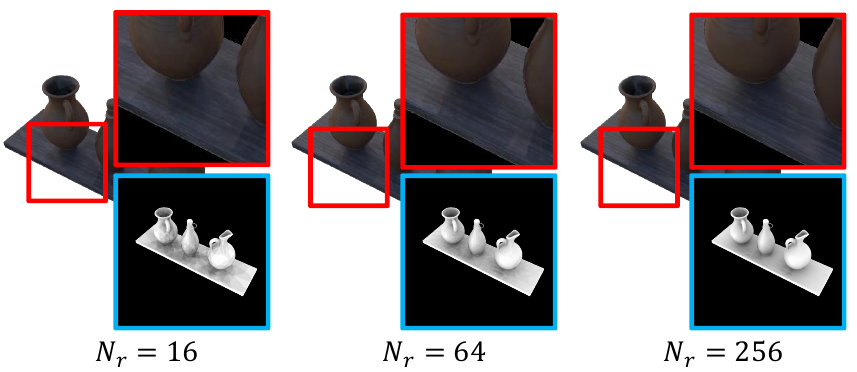}
    \caption{Ablation studies on number of sampled rays $N_\mathrm{r}$.}
    \label{fig:ablation_sample_num}
\end{subfigure}
\vspace{-3mm}
\caption{
Ablation studies on various components of \model{}.
}
\vspace{-8mm}
\label{fig:ablation}
\end{figure}
\begin{table}[t]
    \centering
    \caption{Ablation studies on various components of \model{}.}
    \vspace{-3mm}
    \resizebox{\linewidth}{!}{
        \begin{tabular}{c|c|c|c}
            \toprule
            Method & NVS PSNR $\uparrow$  &  Albedo PSNR $\uparrow$& Relighting PSNR $\uparrow$\\ \hline 
            Detach indirect & 34.21   & 30.29  & 34.22 \\
            w/o indirect (train) & 34.09 &  30.10 & 33.93 \\
            w/o indirect (relight) & - &  - & 33.84 \\
            $N_\mathrm{r}=16$ & 34.01 & 30.21 & 29.46 \\
            $N_\mathrm{r}=64$ & 34.98 & 30.63 & 33.11 \\
            Full & \textbf{35.48} & \textbf{30.81} & \textbf{34.68}  \\
            \bottomrule
        \end{tabular}
    }
\vspace{-4mm}
    \label{tab:ablation}
\end{table}
\subsection{Ablation study}
In \cref{tab:ablation}, we conduct ablation studies on various components of \model{} to validate their effectiveness. We report the average PSNR on NVS, albedo estimation, and relighting across four scenes in Synthetic4Relight dataset~\cite{zhang2022modeling}.

\noindent{\bf Indirect light}
We perform ablation studies to evaluate the impact of indirect light modeling. ``Detach indirect'' indicates that gradients are not backpropagated through ray tracing to optimize indirect radiance. ``w/o indirect (train)'' signifies that the indirect term in \cref{eq:lighting_decompose} is omitted during training, while “w/o indirect (relight)” omits the proposed strategy for computing indirect radiance during relighting. As shown in \cref{fig:ablation_indirect}, detaching the indirect term results in unrealistic indirect radiance, and training without the indirect term leads to inaccurate albedo estimation. We also demonstrate the impact of indirect radiance on relighting results of a composited scene in \cref{fig:teaser} (indirect illumination). Note that the renderings in \cref{fig:teaser} are produced by assigning an additional high metallic to the Gaussian-based ``Ground'', enhancing the visual clarity of inter-reflections.

\noindent{\bf Number of rays}
We also ablate on the number of sampled rays $N_\mathrm{r}$ for rendering equation. We investigated the effects of $N_\mathrm{r}=16$, $N_\mathrm{r}=64$, and $N_\mathrm{r}=256$. As shown in \cref{fig:ablation_sample_num}, more sampled rays lead to more accurate and smooth renderings and ambient occlusion in novel view synthesis.

\section{Conclusion}
In this paper, we introduce inter-reflective Gaussian splatting (\model{}), a novel framework that overcomes the limitations of previous Gaussian-based approaches in accurately modeling visibility and indirect radiance for incident light in inverse rendering. \model{} incorporates the full rendering equation without simplifications and utilizes a differentiable 2D Gaussian ray tracing technique for precise computation of incident radiance at intersection points. 
We further propose an optimization scheme to address the computational demands of Monte Carlo sampling in rendering equation evaluation, and a novel strategy for querying indirect radiance during relighting. 
Extensive experiments on multiple benchmarks validate \model{}’s ability to accurately model complex inter-reflection, establishing it as an effective solution for inverse rendering.

\noindent{\bf Limitation}
Due to the significant computational demands of rendering equation evaluation, we are unable to render at a high frame rate. However, this could be addressed by baking the incident radiance or pre-computing radiance transfer, which we leave as future work.

\section*{Acknowledgments}
This work was supported in part by National Natural Science Foundation of China (Grant No. 62376060).

{
    \small
    \bibliographystyle{ieeenat_fullname}
    \bibliography{main}
}

\clearpage
\setcounter{page}{1}

\twocolumn[{
\renewcommand\twocolumn[1][]{#1}%
\maketitlesupplementary
\begin{center}
\centering
\includegraphics[width=\linewidth]{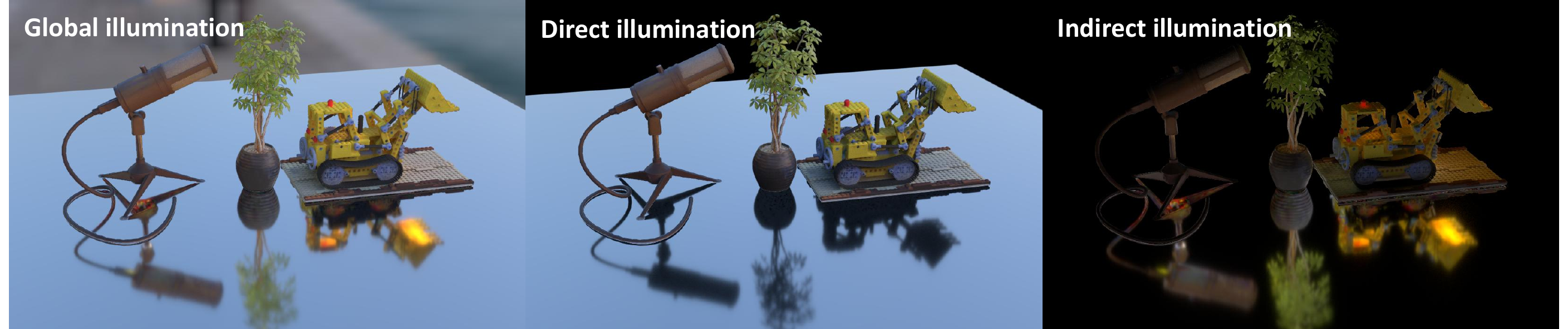}
\captionof{figure}{
Global, direct, and indirect illumination in a Gaussian-based scene using our \model{}.
}
\label{fig:supp_teaser}
\end{center}
}]

\section{Relighting details}

Given a sampled incident direction, we conduct 2D Gaussian ray tracing to obtain the intersected Gaussians along the ray, and aggregate the albedo, roughness, and normal by alpha-blending: $\{\mathcal{A}_r, \mathcal{R}_r, \mathcal{N}_r\} = \sum_{i}\omega_i\{\boldsymbol{a}_i, r_i, \boldsymbol{n}_i\}$, where $\omega_i=\frac{T_\mathrm{i}\alpha_\mathrm{i}}{\sum_\mathrm{i} T_\mathrm{i}\alpha_\mathrm{i}}$. 
Then, to avoid the extensive Monte Carlo sampling, we pre-integrate the cubemap, which allows us to obtain the diffuse $L_\mathrm{d}$ and specular term $L_\mathrm{s}$ for incident direction using only a single query. Specifically, we apply split-sum approximation for the specular term:
\begin{align}\label{eq:split sum}
L_s &\approx \int_{\Omega} f_s(\boldsymbol{\omega}_i, \boldsymbol{\omega}_o) (\boldsymbol{\omega}_i \cdot \mathcal{N}_r) d\boldsymbol{\omega}_i \cdot \notag \\
&\quad \int_{\Omega} L_i(\boldsymbol{\omega}_i) D(\boldsymbol{\omega}_i, \boldsymbol{\omega}_o) (\boldsymbol{\omega}_i \cdot \mathcal{N}_r) d\boldsymbol{\omega}_i,
\end{align}
where the left term depends solely on $(\omega_i \cdot \mathcal{N}_r)$ and roughness $R_r$, this allows the results to be pre-integrated and stored in a 2D lookup texture map. The right term represents the integral of incident radiance, which can also be pre-integrated before rendering. Consequently, the indirect radiance of the incident ray is given as: $L_\mathrm{ind}=L_\mathrm{d} + L_\mathrm{s}$.

\section{Composited scene details}

In \cref{fig:teaser} and \cref{fig:supp_teaser}, we relight a scene composed of four Gaussian-based objects: ``Mic,'' ``Ficus,'' ``Lego,'' and ``Ground.'' The objects ``Mic,'' ``Ficus,'' and ``Lego'' are reconstructed using the proposed \model{} framework, while ``Ground'' is manually designed using a set of parallel 2D Gaussians. To better illustrate the inter-reflective properties of \model{}, we assign an additional metallic property $m$ to each Gaussian. Specifically, we set $m=0$ for ``Mic'', ``Ficus'', and ``Lego'', and $m=1$ for ``Ground''. For efficient rendering, we employ importance sampling with 512 rays distributed using cosine-weighted sampling for the diffuse term and 256 rays distributed using GGX sampling for the specular term. In \cref{fig:supp_teaser}, ``Direct Illumination'' considers only the direct incident radiance, ``Indirect Illumination'' accounts for only the indirect radiance, and ``Global Illumination'' combines both direct and indirect radiance for full rendering.

\section{More results}
\subsection{Results on Synthetic4Relight}
\begin{figure*}[ht]
\centering
\includegraphics[width=\linewidth]{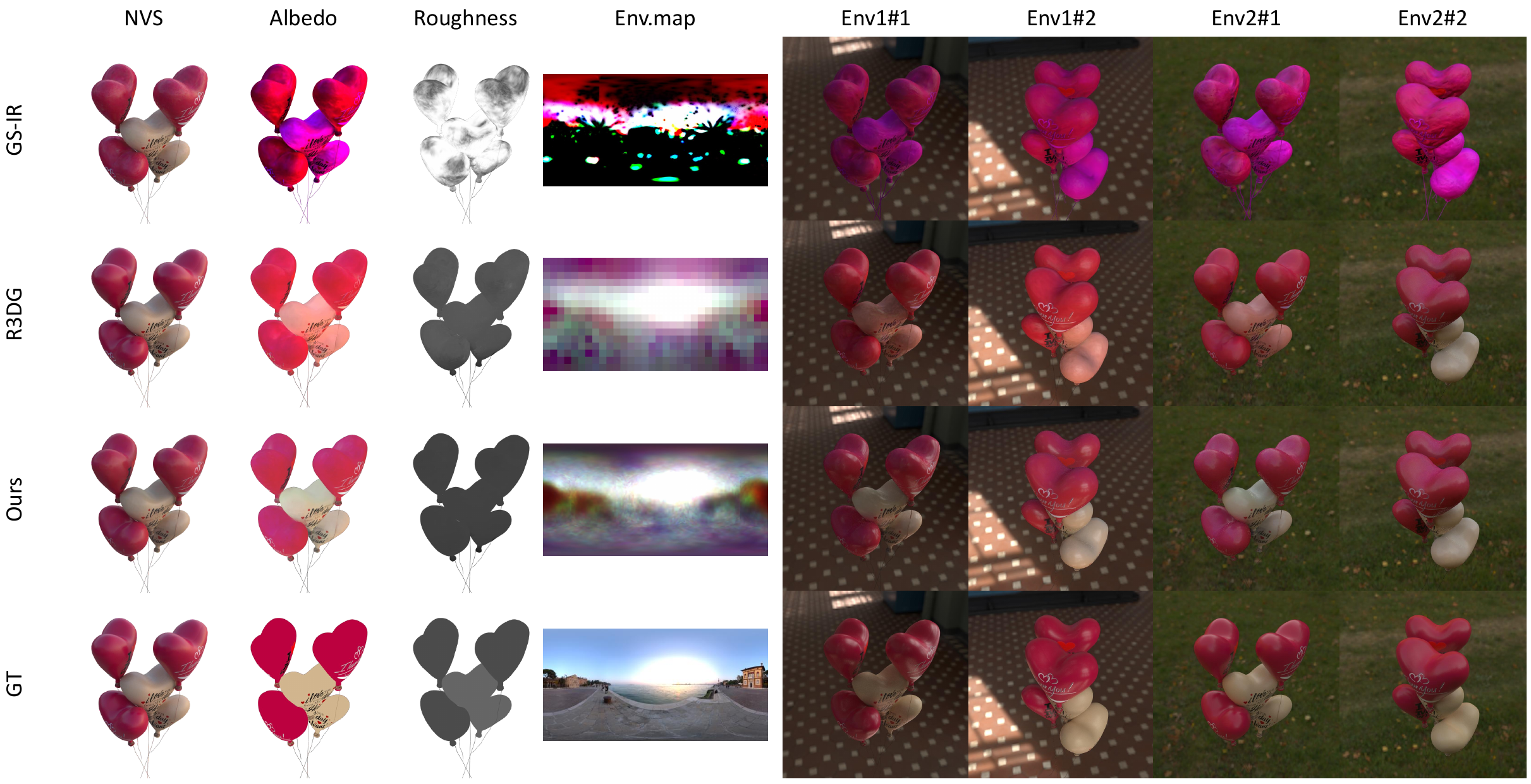}
\caption{
Qualitative comparison of NVS, material and lighting estimation, and relighting results on the Synthetic4Relight dataset~\cite{zhang2022modeling}.
}
\label{fig:supp_exp_syn4_compare_1}
\end{figure*}

\begin{figure*}[ht]
\centering
\includegraphics[width=\linewidth]{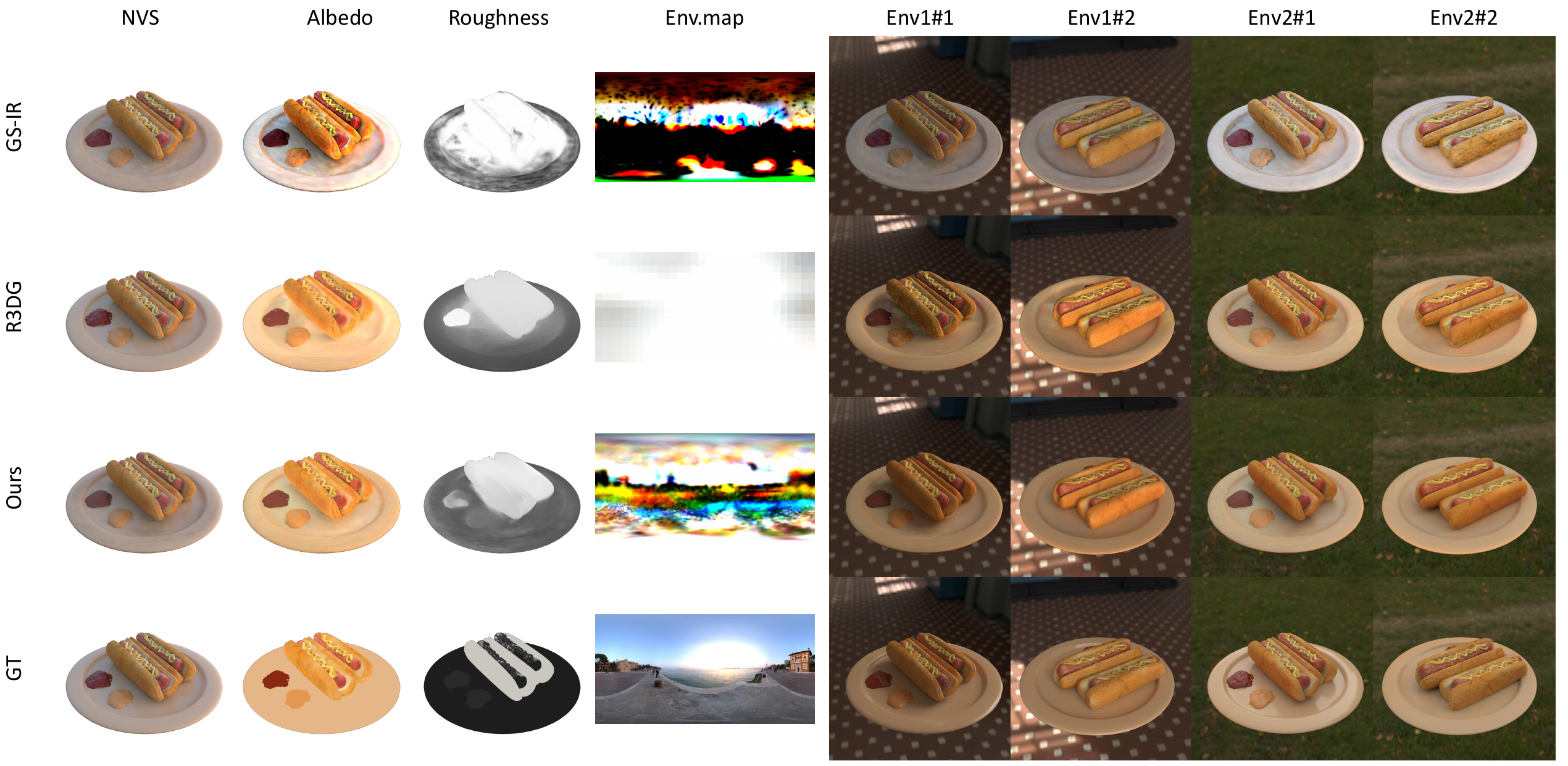}
\caption{
Qualitative comparison of NVS, material and lighting estimation, and relighting results on the Synthetic4Relight dataset~\cite{zhang2022modeling}.
}
\label{fig:supp_exp_syn4_compare_2}
\end{figure*}

\begin{figure*}[ht]
\centering
\includegraphics[width=\linewidth]{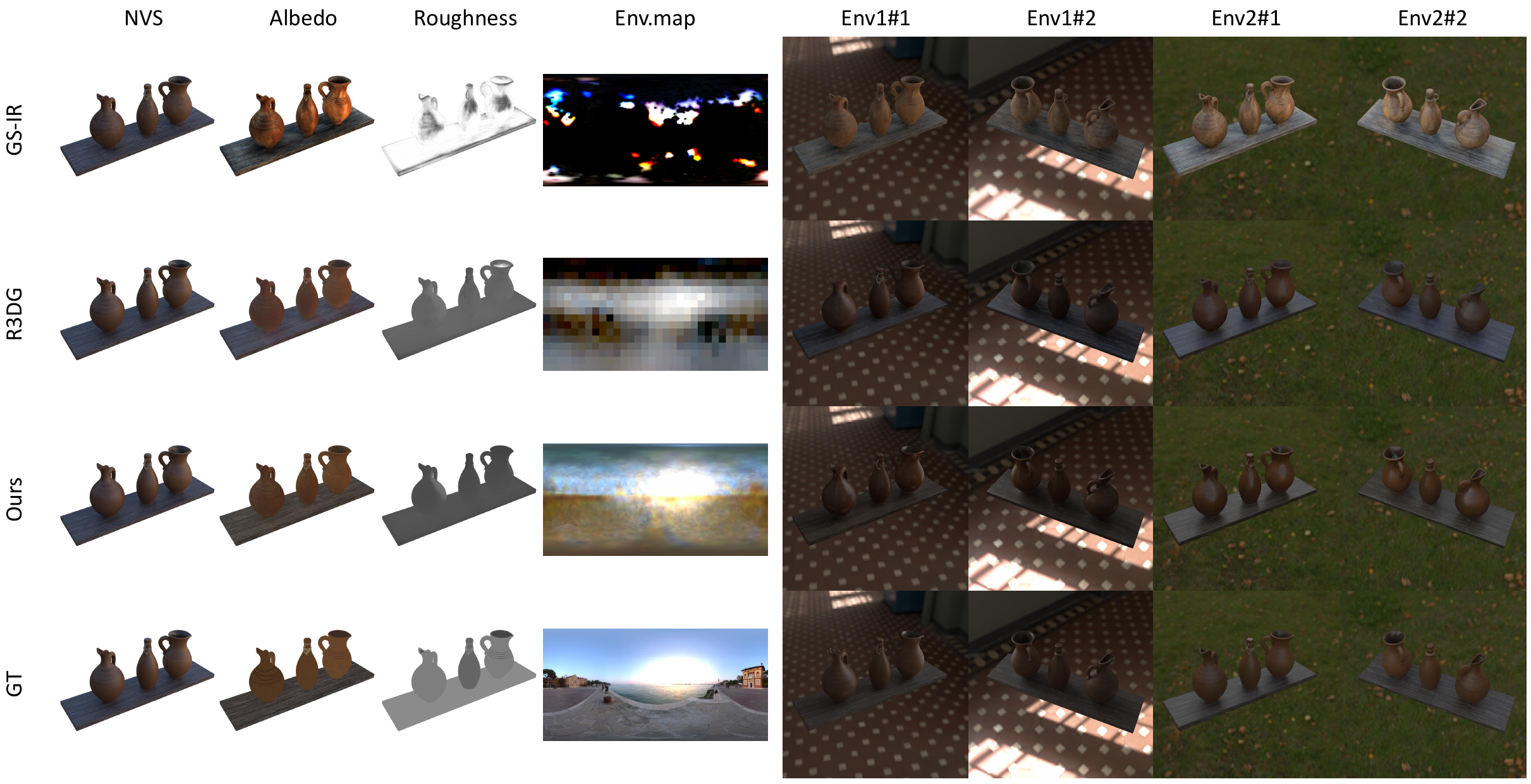}
\caption{
Qualitative comparison of NVS, material and lighting estimation, and relighting results on the Synthetic4Relight dataset~\cite{zhang2022modeling}.
}
\label{fig:supp_exp_syn4_compare_3}
\end{figure*}

\begin{figure*}[ht]
\centering
\includegraphics[width=\linewidth]{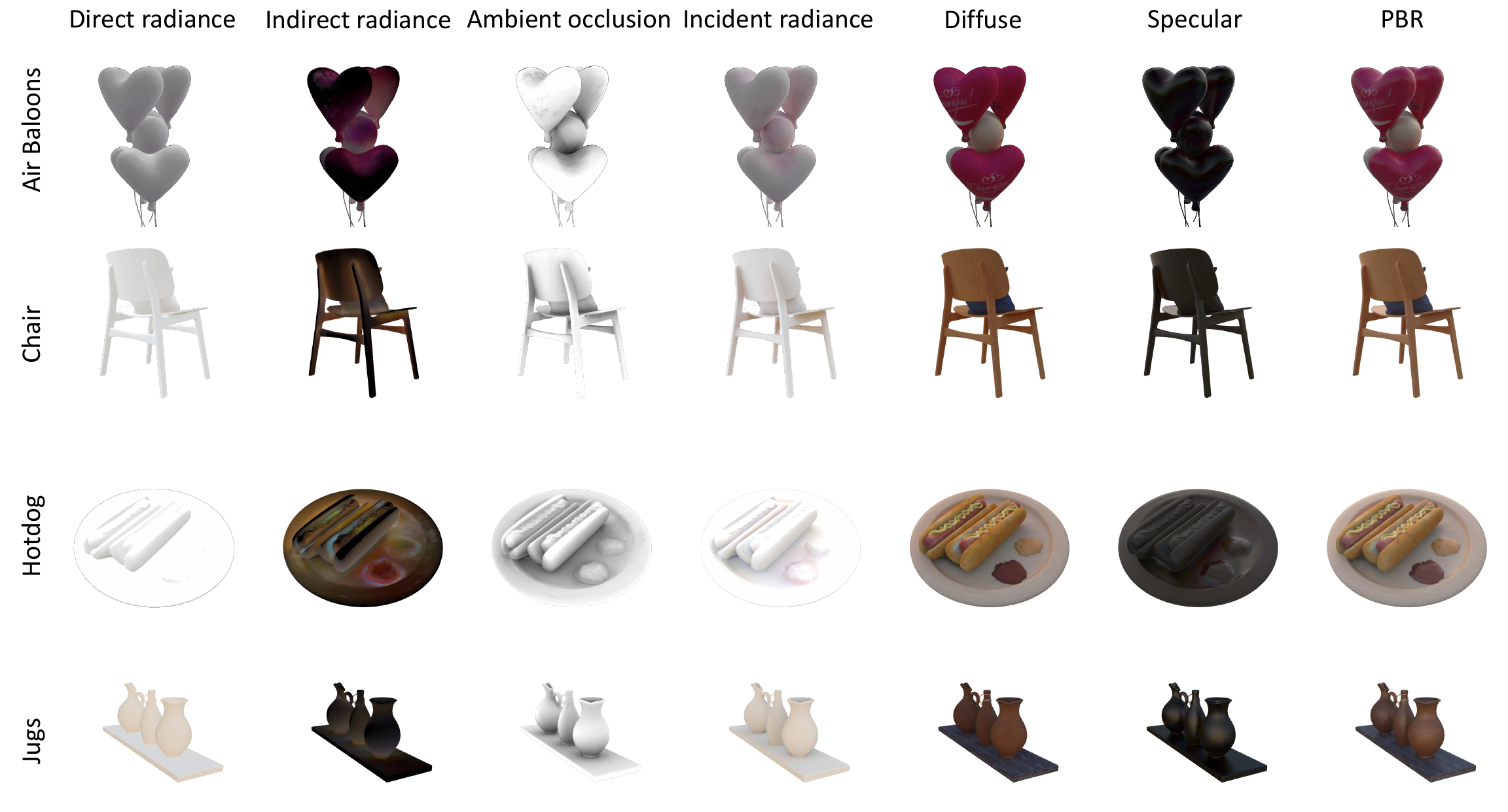}
\caption{
Visualization of estimated components in novel view, including the averaged direct radiance $L_\mathrm{dir}$, averaged indirect radiance $L_\mathrm{ind}$, averaged visibility $V$ (ambient occlusion), averaged incident radiance $L_\mathrm{i}$, diffuse, specular, and final PBR color.
}
\label{fig:supp_exp_syn4_all}
\end{figure*}

\begin{figure*}[ht]
\centering
\includegraphics[width=0.7\linewidth]{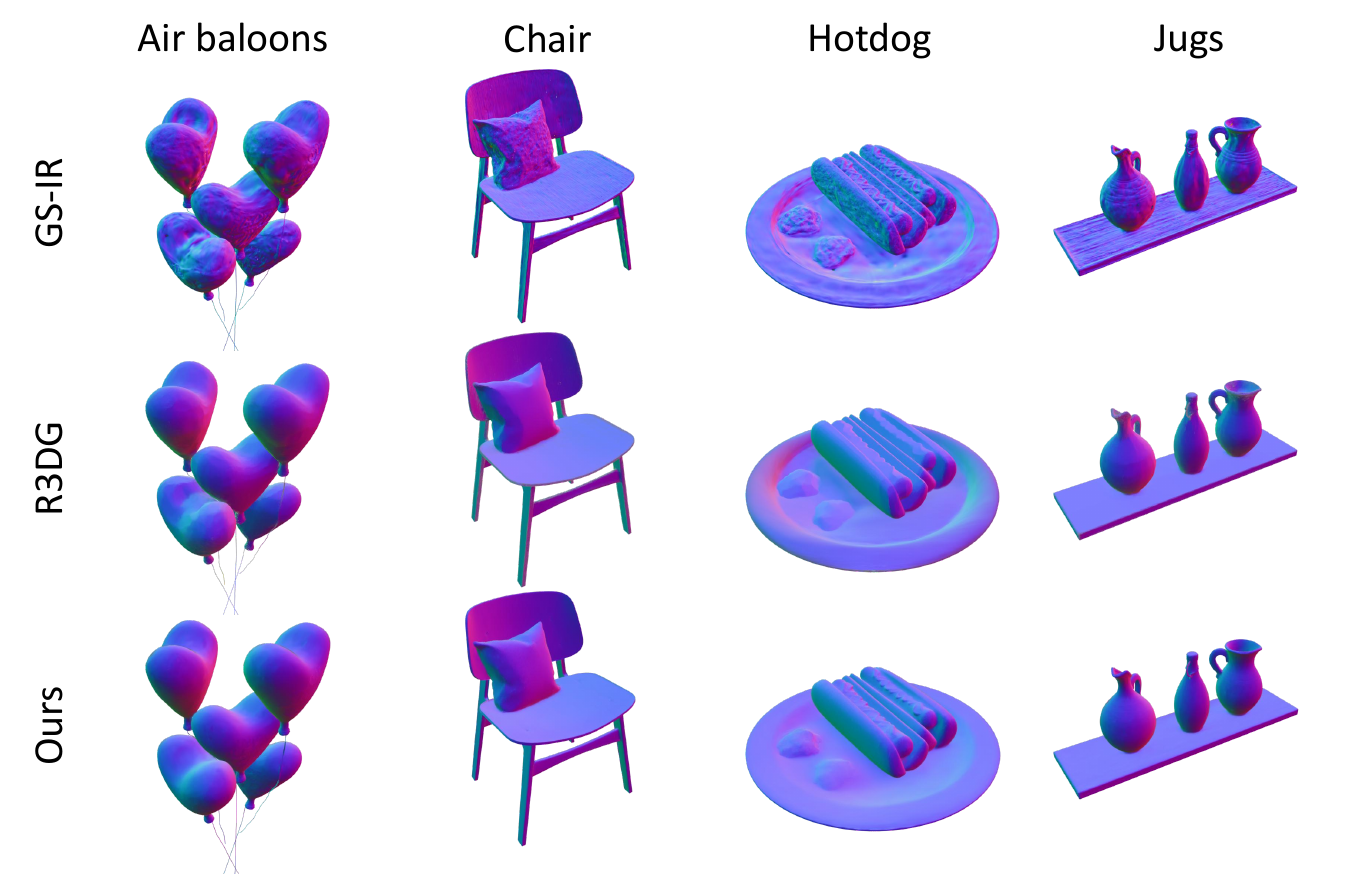}
\caption{
Qualitative comparison of rendered normal maps on the Synthetic4Relight dataset~\cite{zhang2022modeling}. Note that, Synthetic4Relight dataset does not provide ground truth normal maps.
}
\label{fig:supp_exp_syn4_normal_compare}
\end{figure*}

We further provide a qualitative comparison on three additional scenes from the Synthetic4Relight dataset~\cite{zhang2022modeling}, including ``air ballons'' (\cref{fig:supp_exp_syn4_compare_1}), ``hotdog'' (\cref{fig:supp_exp_syn4_compare_2}), and ``jugs'' (\cref{fig:supp_exp_syn4_compare_3}). It is evident that our estimated material properties, environment lighting, and relighted images are the most realistic compared to the competitors, GS-IR~\cite{liang2024gs} and R3DG~\cite{gao2023relightable}. In \cref{fig:supp_exp_syn4_all}, we further visualize the different estimated components in novel view, illustrating \model{}'s ability to capture accurate inter-reflection. Additionally, in \cref{fig:supp_exp_syn4_normal_compare}, we compare the rendered normal maps of each scene with those from GS-IR~\cite{liang2024gs} and R3DG~\cite{gao2023relightable}. Thanks to the accurate geometry provided by 2D Gaussian primitives, our normal maps exhibit superior fidelity.

\begin{figure*}[ht]
\centering
\includegraphics[width=0.7\linewidth]{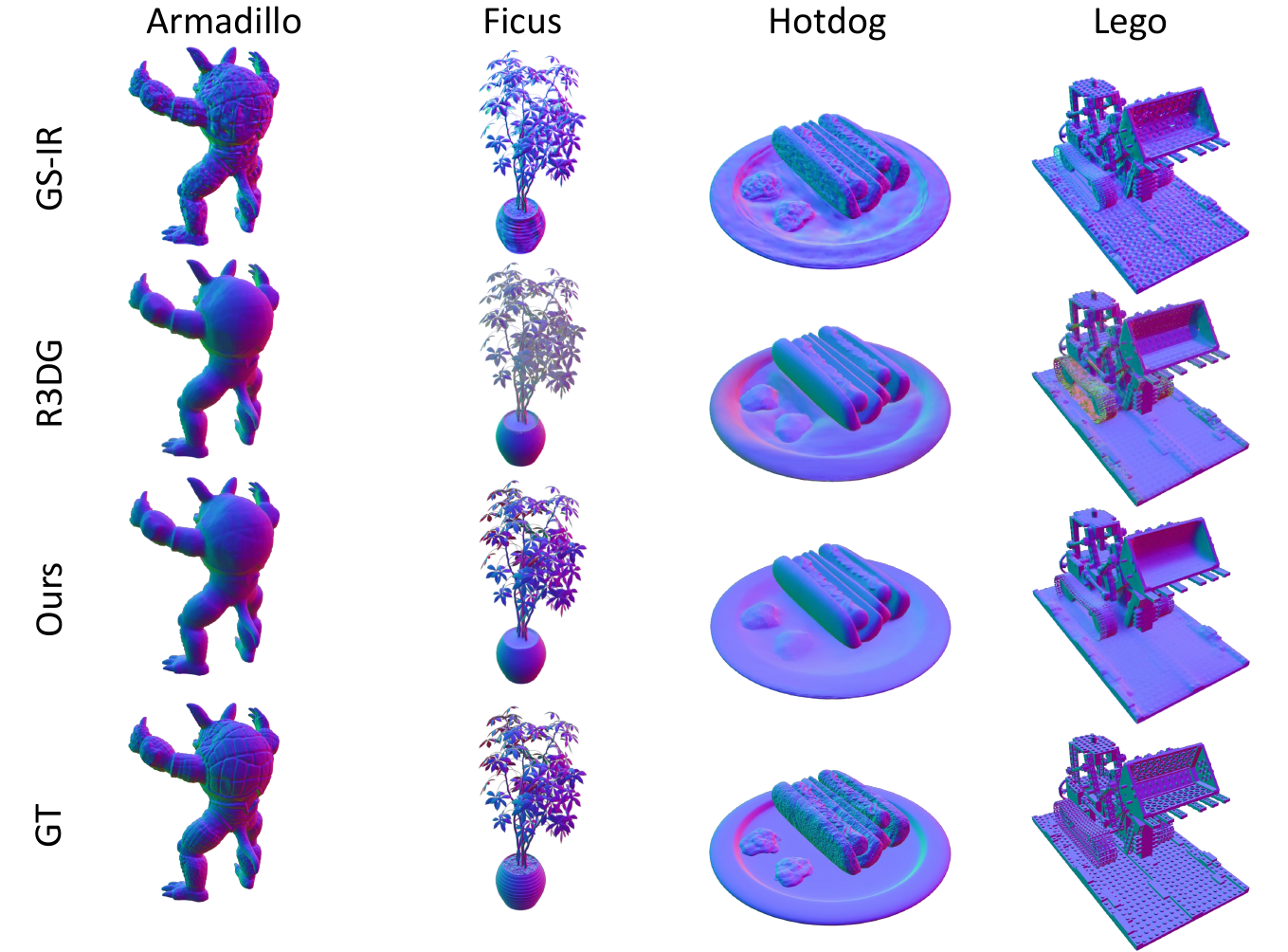}
\caption{
Qualitative comparison of rendered normal maps on the TensoIR dataset~\cite{jin2023tensoir}.
}
\label{fig:supp_exp_tensoir_normal_compare}
\end{figure*}
\begin{figure*}[ht]
\centering
\includegraphics[width=0.7\linewidth]{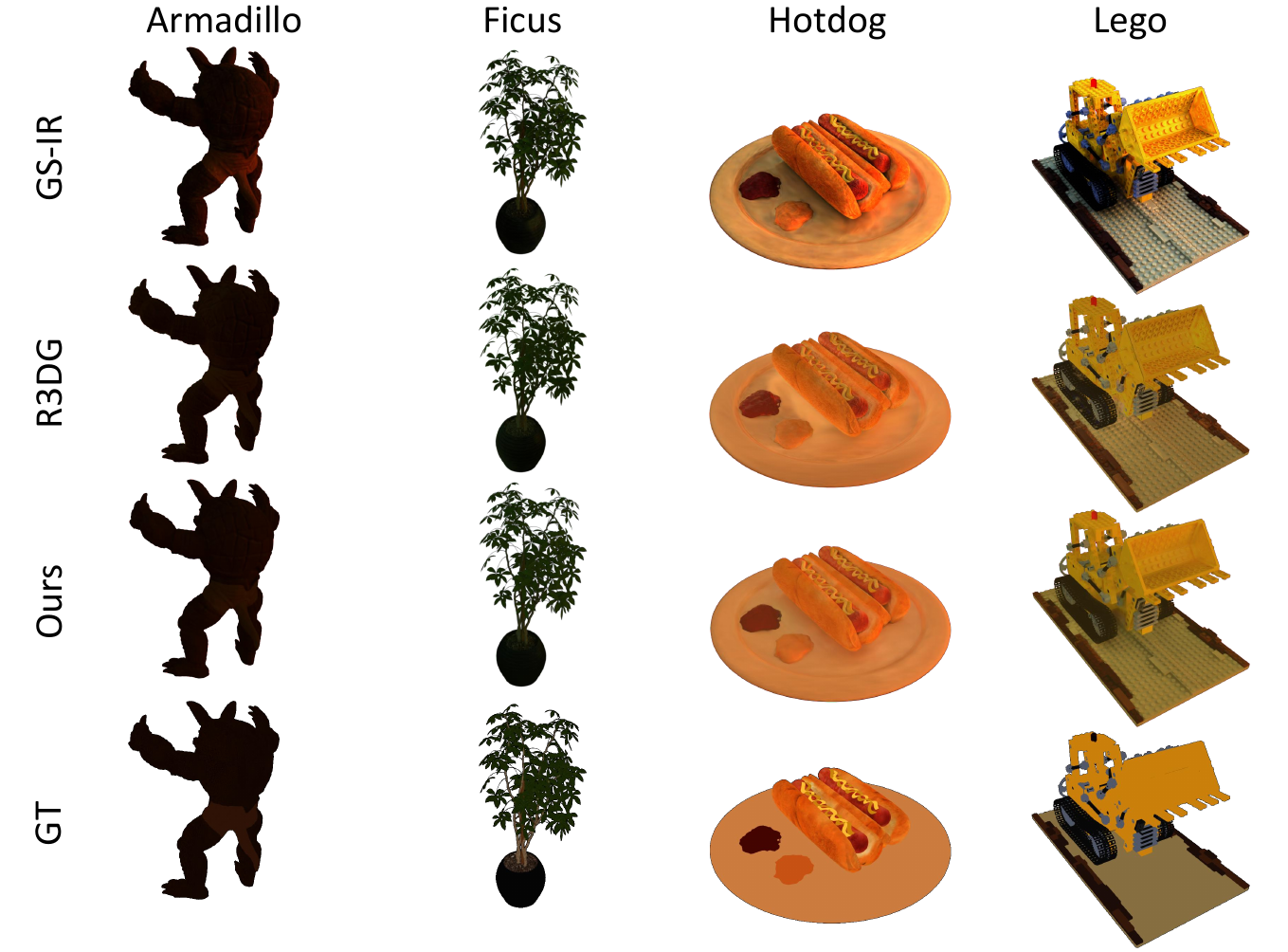}
\caption{
Qualitative comparison of estimated albedo maps on the TensoIR dataset~\cite{jin2023tensoir}. Note that, we scale each RGB channel by a global scalar.
}
\label{fig:supp_exp_tensoir_albedo_compare}
\end{figure*}

\subsection{Results on TensoIR}
We present a qualitative comparison of the rendered normal maps and estimated albedo maps in \cref{fig:supp_exp_tensoir_normal_compare} and \cref{fig:supp_exp_tensoir_albedo_compare}, respectively. \model{} produces smooth normal maps, and clean albedo maps with minimal shadow artifacts, attributed to the accurate modeling of incident radiance through 2D Gaussian ray tracing.

\end{document}